%% file: main.tex
\documentclass[runningheads]{llncs}

\usepackage[mobile]{eccv}

\usepackage{eccvabbrv}

\usepackage{graphicx}
\usepackage{booktabs}
\usepackage{makecell} 
\usepackage{microtype} 
\usepackage{multirow}
\usepackage{subcaption}

\usepackage[accsupp]{axessibility}  

\usepackage[
    colorlinks=true,
    linkcolor=magenta,
    urlcolor=magenta,
    citecolor=eccvblue
]{hyperref}

\usepackage{orcidlink}

\begin{document}

\title{HybridSim: A Physics–Learning Hybrid Digital Twin for mmWave Human Sensing}

\titlerunning{HybridSim}

\author{
\centering
Weitao Xiong\inst{1,2} \quad
Tianyu Liu\inst{2} \quad
Peng Li\inst{2} \quad
Kok Chung Chua\inst{1} \quad\\[0.25em]
Toa Chean Khim\inst{1} \quad
Pu Wang\inst{3} \quad
Hongfei Xue\inst{3\dagger}\\[0.5em]
\small \url{https://weitao-xiong.github.io/HybridSim/}\vspace{-0.5em}
}

\authorrunning{W. Xiong et al.}

\institute{
\centering
$^{1}$Xiamen University Malaysia \quad \\[0.25em]
$^{2}$The Hong Kong University of Science and Technology \quad \\[0.25em]
$^{3}$University of North Carolina at Charlotte
\\[0.5em]
$^{\dagger}$Corresponding Author
}

\maketitle
\vspace{-2em}
\begin{figure}[h]
    \centering
    \includegraphics[width=1\linewidth]{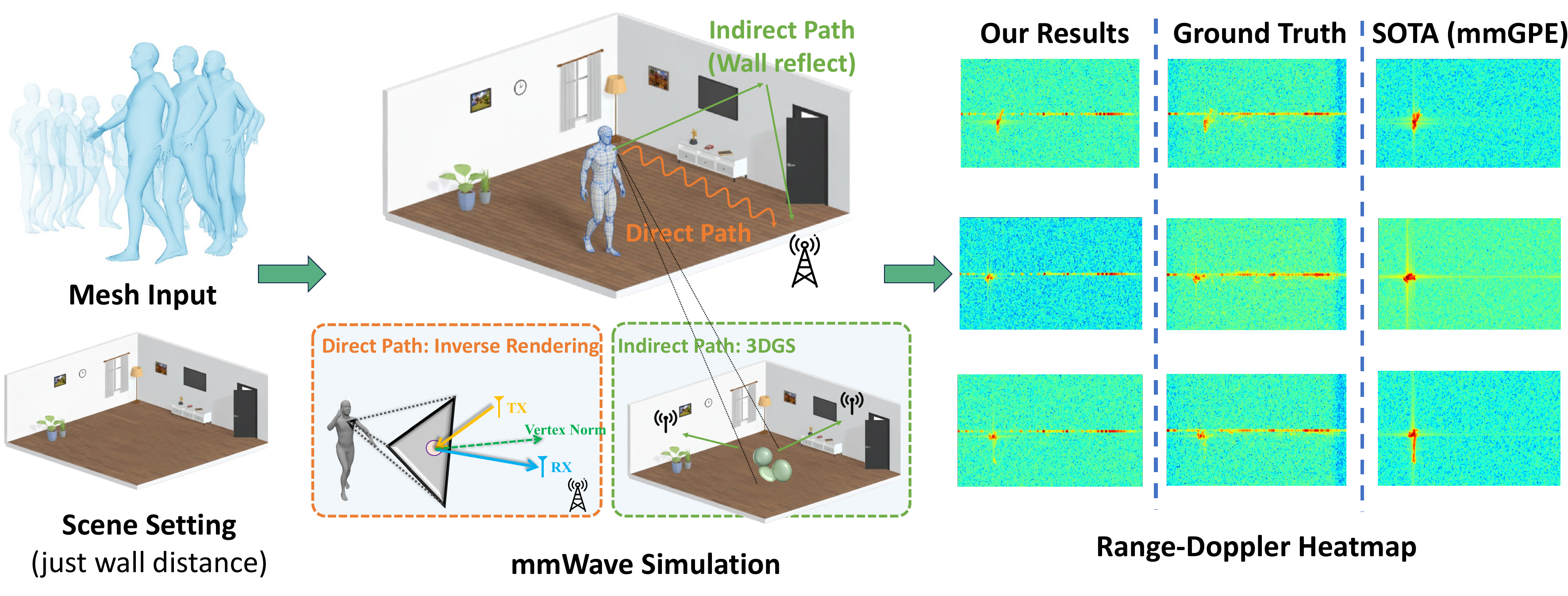}
    \caption{\textbf{HybridSim: Dynamic Human Motion mmWave Simulation via Physics–Learning \& Decoupled Hybrid Rendering.} HybridSim synthesizes realistic mmWave signals from dynamic human meshes and indoor scenes by decoupling propagation into direct (inverse rendering-based) and indirect (3D Gaussian Splatting-based) paths. Compared to existing simulators, HybridSim preserves fine-grained kinematic details in Range-Doppler heatmaps, showing high fidelity to the physical ground truth measurements.}
    \label{fig:teaser}
\end{figure}

\vspace{-3em}
\begin{abstract}
High-fidelity simulation of mmWave radar signals for dynamic human motion is valuable for developing radar-based human sensing models; yet collecting accurately labeled measurements for a specific deployment site remains expensive. We present HybridSim, a physics–learning hybrid simulator that synthesizes mmWave radar signals from dynamic human meshes under a fixed indoor room configuration, explicitly decoupling propagation into two components. To parameterize the human subject, we use a tri-plane representation to extract human features and a Graph Convolutional Network to stabilize optimization and mitigate gradient instability. The direct signal path is modeled via an inverse-rendering formulation with a microfacet BRDF to capture primary surface reflections. In parallel, the indirect path is approximated by combining 3D Gaussian Splatting with a virtual-receiver geometry to fit and reproduce site-specific multipath interference patterns, achieving substantially lower computational cost than explicit full ray tracing. Experiments in a fixed-room setting show improved agreement with a physically based reference and consistent gains on downstream radar-based human sensing tasks when using HybridSim for site-specific data augmentation.
\end{abstract}

\section{Introduction}
\label{sec:intro}
Millimeter-wave radar provides a robust and cost-effective alternative to vision-based human sensing systems due to its resilience to adverse lighting and preservation of privacy. Developing robust deep learning models for radar perception requires extensive training datasets encompassing diverse human activities. Collecting and annotating such data in physical environments is highly resource-intensive. Consequently, synthesizing high-fidelity radar signals through physical simulation serves as a practical solution to scale training datasets and bridge the domain gap between synthetic and real-world measurements~\cite{10.1145/3458864.3467679,10.1145/3560905.3568545}.

Existing mmWave simulation methods face distinct challenges when rendering continuous dynamic human motion. Unlike high-density LiDAR, mmWave yield exceedingly sparse spatial measurements. Consequently, radar-based human sensing relies heavily on time-varying micro-Doppler signatures, rather than dense spatial geometry, to infer action. Physics-based ray tracing frameworks provide accurate electromagnetic propagation models but suffer from exponential computational complexity when evaluating higher-order multipath interactions~\cite{7152831,10.1145/3625687.3625798}. Furthermore, some ray tracing methods highly demand precise scanned geometries of the environment, making them inconvenient for ubiquitous deployment~\cite{10556262}. Conversely, recent neural representations adapt radiance fields and Gaussian splatting to the radio frequency domain~\cite{10.1145/3570361.3592527,11355734}. While these approaches excel at static site-specific channel modeling, they inherently entangle the primary surface scattering of the dynamic human body with complex environmental multipath reflections. This entanglement degrades the synthesis of time-varying micro-Doppler signatures, which are essential for identifying articulated kinematics in single-sensor setups. Other data-driven generation methods and physical augmenters address these efficiency issues but often compromise on coherent multipath interference and physical fidelity~\cite{10.1145/3560905.3568542,10.1145/3570361.3613302}.

To resolve these limitations, we present HybridSim, a hybrid physics and learning simulator designed to synthesize realistic intermediate-frequency millimeter-wave signals from dynamic human meshes in static indoor environments. We decouple the complex electromagnetic propagation into two distinct and interpretable components. For the direct signal path, we employ an inverse rendering formulation utilizing a microfacet bidirectional reflectance distribution function to accurately capture sharp primary reflections from the subject and the room boundaries. Simultaneously, we model the indirect path by integrating geometrically constrained 3D Gaussian splatting (3DGS)~\cite{10.1145/3592433} with proxy virtual receivers positioned on the room boundaries. This formulation functions as an efficient multi-scatterer surrogate that learns and reconstructs the site-specific multipath interference patterns at a significantly lower computational cost than explicit multi-bounce ray tracing.

By explicitly disentangling the direct surface scattering from the indirect environmental multipath effects, our framework preserves fine-grained kinematic details within the generated Range-Doppler heatmaps. Extensive evaluations confirm that the synthesized signals maintain high physical fidelity to real-world measurements. Furthermore, using the simulated data for data augmentation significantly improves the classification accuracy of downstream radar-based human activity recognition models under cross-subject and sim-to-real evaluation protocols. The primary contributions of this work include a novel decoupled rendering architecture for efficient dynamic radar synthesis, a unified multi-scatterer method for modeling complex multipath effects, and empirical validation of the simulator in enhancing practical human sensing applications.

\section{Related Works}
\label{sec:related_works}
\subsection{RF Synthesis}  
While mmWave systems like mmMesh~\cite{10.1145/3458864.3467679} and M$^{4}$esh~\cite{10.1145/3560905.3568545} offer robust, privacy-preserving human pose estimation, their generalization to unseen activities is often bottlenecked by the scarcity of comprehensive real-world training data. 

To handle these limitations, many approaches have been developed. Data-driven approaches like SynMotion~\cite{10.1145/3560905.3568542} generate radar signals directly from vision-based skeletal data, whereas methods like mmGPE~\cite{10.1145/3570361.3613302} employ physical simulation augmenters to mimic body reflections.

\textbf{Ray-Tracing RF Simulation.} Physics-based simulations rely on geometrical optics and ray tracing to approximate electromagnetic propagation. Foundational work by Yun and Iskander~\cite{7152831} provide a way to
modeling these multi-path effects, offering a reasonable explanation
of how signals will interact with the environment. And Sch\"u{\ss}ler et al.~\cite{9533181} extended for large MIMO arrays in automotive contexts. RF-Genesis~\cite{10.1145/3625687.3625798} integrated ray tracing with diffusion models to scale vision-to-RF data generation across diverse environments. While these ray-tracing pipelines achieve high physical fidelity, their computational cost scales exponentially with higher-order multipath interactions, rendering them inefficient for continuous dynamic motion synthesis.

\textbf{Differentiable RF Rendering.} To enable parameter estimation and end-to-end optimization, recent efforts have introduced differentiable RF rendering. Hofmann et al.~\cite{10892639} adapted inverse rendering to near-field radar to reconstruct material properties from measurements. Similarly, InverTwin~\cite{chen2025invertwinsolvinginverseproblems} targets digital twin construction by addressing local non-convexities and boundary discontinuities in RF simulations. However, these methods are primarily designed for inverse problem solving rather than functioning as efficient forward simulators for highly articulated human dynamics.

\textbf{Neural RF Representations.} The inflexibility of monolithic simulators has motivated the development of neural RF representations. Frameworks such as RFCanvas~\cite{10.1145/3666025.3699351} and RFScape~\cite{11092981} fuse visual priors with neural distance fields to build editable channel models. Concurrently, radiance field and 3DGS~\cite{10.1145/3503250, 10.1145/3592433} have been adapted for RF propagation. Pioneering this shift, NeRF$^2$~\cite{10.1145/3570361.3592527} translates optical neural radiance fields to the RF domain by employing a complex-valued multilayer perceptron to model both amplitude and phase alongside a Friis-based ray tracing model. RF-3DGS~\cite{11355734} reconstructs radio radiance fields from sparse measurements, and GSRF~\cite{yang2025gsrfcomplexvalued3dgaussian} introduces complex-valued Gaussian primitives with directional bases to model amplitude and phase. 

While these methods excel at static channel modeling, dynamic human sensing imposes distinct requirements. The simulated signal must preserve the time-varying micro-Doppler signatures of articulated motion while maintaining fidelity to coherent multipath interference. Existing neural methods often entangle these phenomena, making it difficult to isolate the moving target from static environment reflections. Furthermore, some methods rely on distributed multi-sensor setups and will suffer significant performance degradation in practical single-sensor scenarios. HybridSim addresses these gaps by explicitly decoupling the Bidirectional Reflectance Distribution Function (BRDF) based primary scattering from the 3DGS-based multipath reflections.

\subsection{Human Modeling} 
Parametric human body models, such as SCAPE~\cite{10.1145/3596711.3596797}, SMPL~\cite{10.1145/2816795.2818013}, and SMPL-X~\cite{8953319} represent 3D humans using compact shape and pose parameters applied to a fixed-topology mesh, enabling efficient animation and optimization. In HybridSim, we use SMPL as the mesh template to get dynamic human poses.

\begin{figure}[t]
    \centering
    \includegraphics[width=1\linewidth]{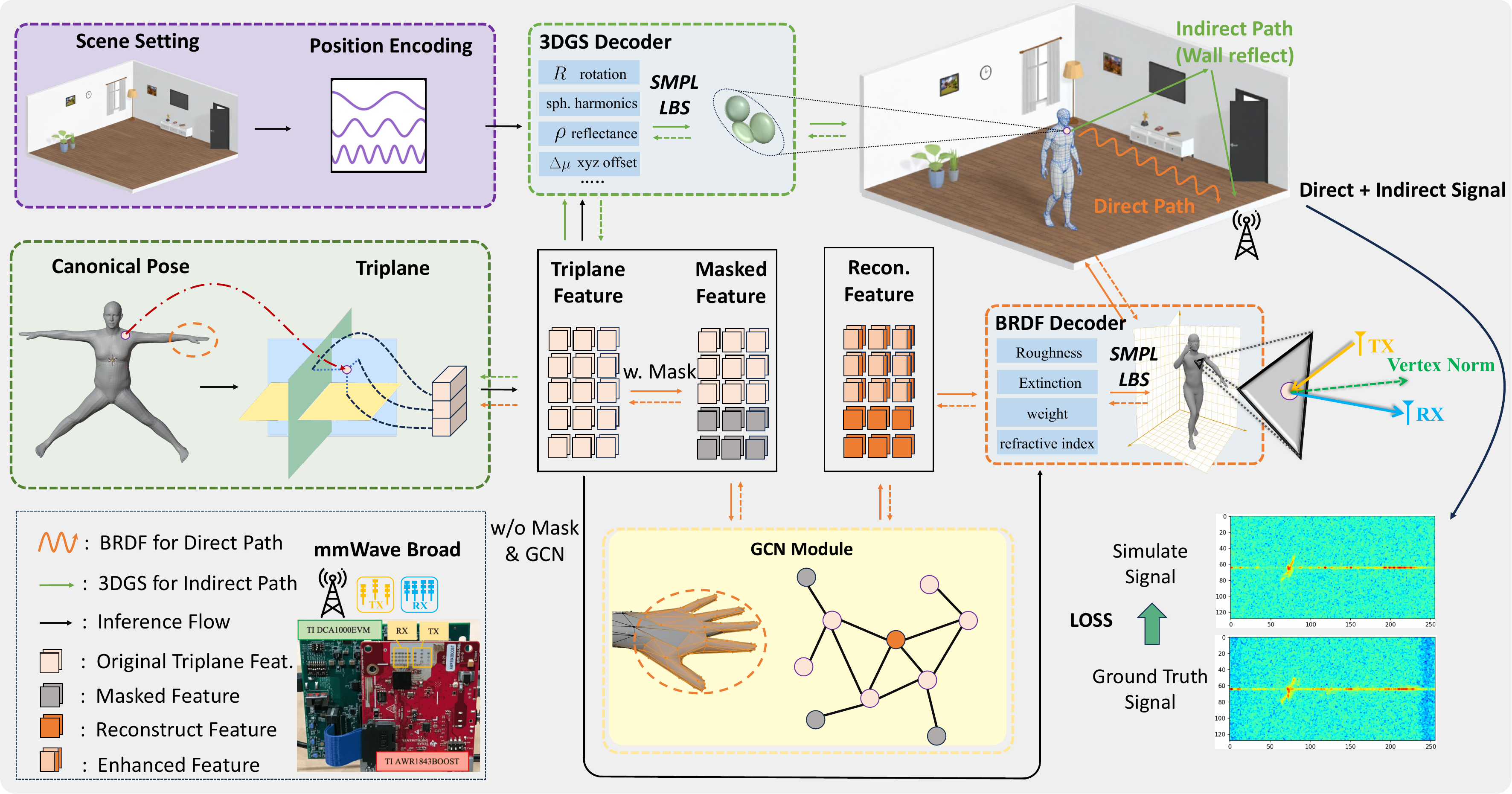}
    \caption{\textbf{Pipeline of our framework.} We extract human features from a canonical pose using a tri-plane representation and obtain scene features through positional encoding. To model the direct path, the human features are masked and reconstructed via a Graph Convolutional Network to generate spatially regularized embeddings. These enhanced features are processed by a BRDF Decoder to predict material properties including roughness and extinction. Concurrently, for the indirect path, the unmasked Triplane features are fed into a 3DGS Decoder to predict parameters such as rotation, spherical harmonics, reflectance, scaling, and vertex offsets. Both sets of decoded parameters are transformed into the dynamic target pose space using Linear Blend Skinning. Finally, we synthesize the direct and indirect signals, and compute the loss against the ground truth signal.}
    \label{fig:pipeline}
\end{figure}

\section{Method}
\textbf{Overview.}
HybridSim is a hybrid physics-learning framework that takes dynamic human mesh sequences and static room configurations as inputs and outputs high-fidelity intermediate-frequency (IF) mmWave radar signals. As shown in Fig.~\ref{fig:pipeline}, we achieve this synthesis by decoupling complex electromagnetic propagation into two interpretable components: a primary direct path and a multipath indirect reflection path.

Specifically, Sec.~\ref{sec:neural_feature} introduces neural feature representations for the human subject and room boundaries alongside dynamic visibility computation. Next, Sec.~\ref{sec:direct_path} models the direct path using microfacet inverse rendering to capture sharp primary reflections. Sec.~\ref{sec:indirect_path} then details our unified multi-scatterer surrogate model for the indirect path, which approximates complex environmental multipath effects via 3DGS and proxy boundary-geometry. Finally, Sec.~\ref{sec:loss} outlines the Range-Doppler domain loss function for end-to-end optimization.

\subsection{Neural Feature Representation and Visibility}
\label{sec:neural_feature}
To effectively model the complex indoor electromagnetic environment, we parameterize the dynamic human subject and the static scene using distinct neural representations. We extract spatial features for the human target from a canonical SMPL mesh using a tri-plane representation~\cite{9880428, 10655076}. Mapping the canonical vertices into these three orthogonal feature planes yields a continuous representation that remains fundamentally invariant to articulated poses. 

For the static room components, rather than relying on complex scanned geometries, we parametrically construct the environmental boundaries using scalar distance configurations relative to the radar transceiver. Based on these spatial limits, we generate a synthetic, grid-based point cloud for each planar surface, encompassing the walls, floor, and ceiling. Static interior objects such as furniture do not require explicit geometric modeling. Instead, their complex electromagnetic scattering effects are implicitly absorbed and represented by the learnable neural parameters distributed across these proxy boundaries. Then we apply periodic positional encoding to these normalized coordinates to yield the scene feature embeddings, equipping the network to learn high-frequency spatial variations across the simulated environment.


A physically accurate radar simulation inherently requires rigorous handling of self-occlusions and mutual line-of-sight blockages between the subject and the environment. To achieve this, we dynamically determine the visibility of all scatterers using a Hidden Point Removal (HPR) algorithm~\cite{10.1016/j.cag.2010.03.002}. The detailed implementation of visibility computation is in Supplementary Material Sec.~\ref{sec:supp_visibility}.

\subsection{Direct Path Simulation}
\label{sec:direct_path}
The direct signal path strictly evaluates the single-bounce (i.e., line-of-sight scattering) radar reflections returning directly to the receiver from the human body and the static room boundaries. By defining this path purely as 1-bounce propagation, we decouple the complex radar signal into a learned amplitude component and a physically derived phase component.

To determine the scattering amplitude, we first pass the extracted neural features through a Graph Convolutional Network (GCN)~\cite{10.1007/s11063-020-10404-7} with stochastic masking, which emulates partial visibility and encourages robust feature recovery. At mmWave frequencies, the human body surface is well-approximated as a piecewise-smooth electromagnetic material field; treating each mesh vertex as an independent scatterer ignores this spatial coherence and can induce non-physical, high-frequency variations in the inferred material parameters. The GCN introduces a soft structural prior via Laplacian-like propagation on the mesh graph, encouraging locally consistent latent material embeddings and improving optimization stability under the high sensitivity of BRDF parameters. In particular, the smoothing effect mitigates gradient concentration on a small subset of vertices, which otherwise can dominate updates and lead to transient memory spikes in the inverse-rendering stage during training. Accordingly, we use the GCN strictly as a training-time regularizer (randomly enabled during training together with stochastic masking) to stabilize convergence and improve generalization to partial visibility. During inference, we directly decode the converged latent features without additional graph propagation, since further message passing may introduce unnecessary spatial smoothing that can suppress legitimate high-frequency scattering changes, while the continuity prior has already been absorbed into the learned latent field through training. Additional analyses and ablations are provided in Supplementary Material Sec.~\ref{sec:gcn_ablation}.



The human features and encoded scene features are fed into BRDF decoders to predict physical material parameters, including surface roughness, a metallic blending factor, and the complex refractive index. Together with the surface normals of the posed SMPL mesh and static wall patches, these parameters define a unified set of direct-path scatterers $\mathcal{P}_{dir}$. For each scatterer $p \in \mathcal{P}_{dir}$ at time $t$, we model the scattering amplitude $L_p$ using BRDF formulation with complex Fresnel equations \cite{10892639}. More details are in Supplementary Material Sec.~\ref{sec:brdf_details}.

While the network predicts the amplitude, the phase is fully determined by propagation physics through the total delay

\begin{equation}
    \tau_p(t) = \frac{\|\mathbf{x}_{\text{tx}} - \mathbf{x}_p(t) \| + \|\mathbf{x}_p(t) - \mathbf{x}_{rx} \|}{c}, \label{eq:delay}
\end{equation}

where $c$ represents the speed of light, $\mathbf{x}_{\text{tx}}$ is the transmitter location, and $\mathbf{x}_{rx}$ is the receiver location. The phase term within the complex signal formulation is driven by the carrier frequency $f_0$, the chirp slope $S$, and this propagation delay.

By combining the learned amplitude and the physical phase, the final direct mmWave signal is synthesized as the coherent summation of contributions from all globally visible scatterers:

\begin{equation}
    Sim_d(t) = \sum_{p \in \text{vis}} \frac{L_p \cdot \exp \left( j 2 \pi \left( f_0 \tau_p(t) + S t \tau_p(t) \right) \right)}{\|\mathbf{x}_{\text{tx}} - \mathbf{x}_p(t) \| \cdot \|\mathbf{x}_p(t) - \mathbf{x}_{rx} \|}.
    \label{eq:sim_d}
\end{equation}

In this equation, $Sim_d(t)$ denotes the aggregated signal at the receiver, and the summation evaluates over the unified subset of visible scattering centers $p \in \text{vis}$ (determined dynamically via HPR) from both the dynamic human mesh and the static room boundaries.  Following the equation in mmGPE~\cite{10.1145/3570361.3613302}, we omit the negligible residual video phase (RVP) term as well. A fundamental physical distinction between radar signal synthesis and standard optical rendering is that while conventional Bidirectional Reflectance Distribution Function and 3DGS pipelines project 3D properties onto a high-resolution 2D image plane for per-pixel shading and blending, a single radar receiving antenna functions effectively as a spatial point sink (\ie, a single-pixel sensor). Consequently, rather than performing spatial rasterization, the receiver coherently superimposes the complex electromagnetic waves from all visible scattering centers into a unified time-domain sequence. Ultimately, the denominator explicitly accounts for physical free-space attenuation, scaling the signal amplitude inversely with the geometric distances between the transmitter, the scattering elements, and the receiver.

\begin{figure}[t]
    \centering
    \begin{subfigure}[t]{0.25\linewidth}
        \centering
        \includegraphics[height=3.15cm,keepaspectratio]{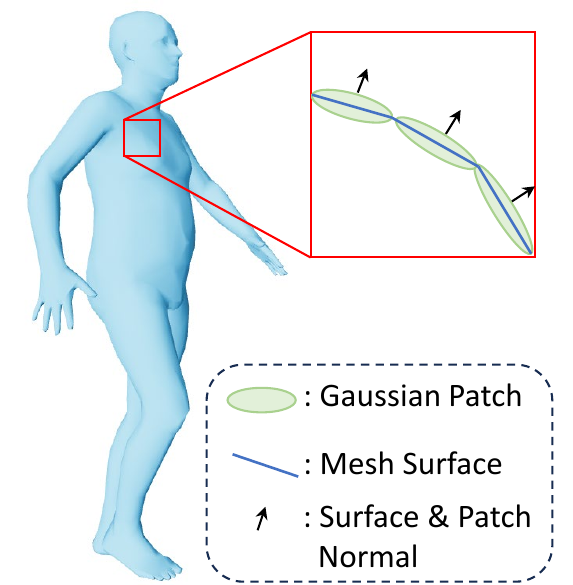}
        \caption{}
        \label{fig:left}
    \end{subfigure}
    \hfill
    \begin{subfigure}[t]{0.74\linewidth}
        \centering
        \includegraphics[height=3.15cm,keepaspectratio]{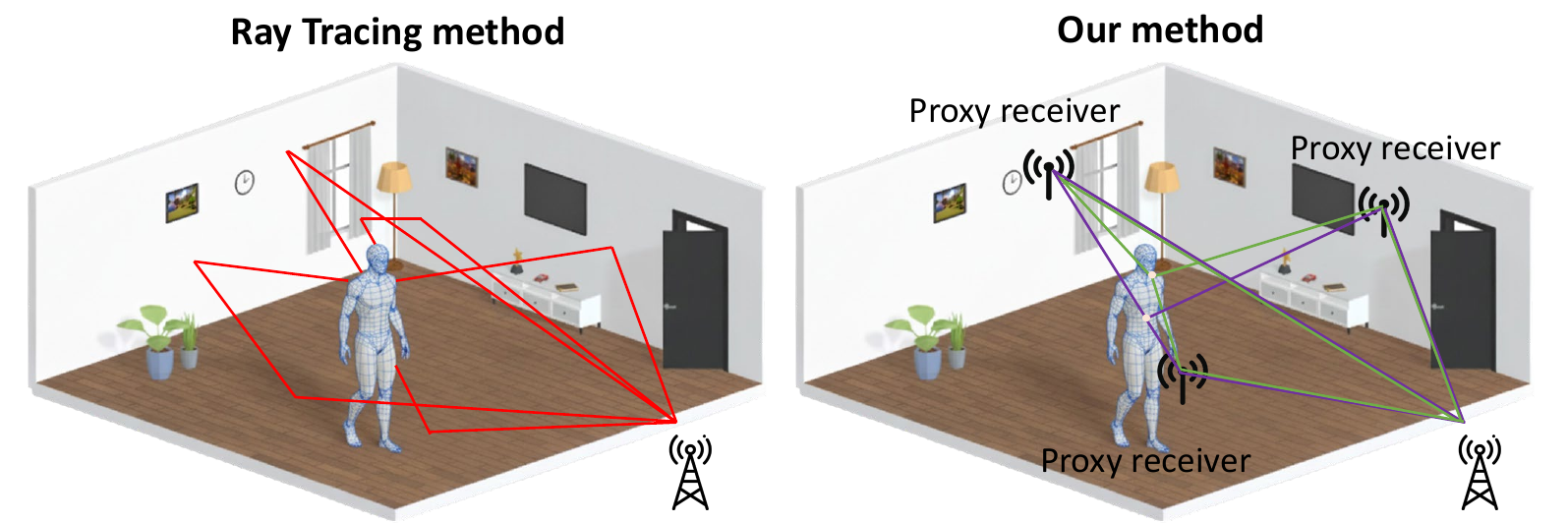}
        \caption{}
        \label{fig:right}
    \end{subfigure}
    
    \caption{\textbf{Illustration of the indirect path simulation using surface-aligned Gaussian patches and proxy wall-geometry.} (\textbf{a}) The dynamic human mesh is augmented with geometrically constrained Gaussian primitives. To prevent unconstrained volumetric inflation, these scattering elements are flattened into 2D surface patches, with their principal axes aligned to the local tangent plane and compressed along the surface normal. (\textbf{b}) Comparison between conventional ray tracing and the proposed proxy wall-geometry approach. While traditional multi-bounce ray tracing (left, red paths) entails a continuous and optimization-unfriendly process, the proposed framework (right) discretizes the multipath environment by instantiating virtual wall receivers on the room boundaries. This formulation simplifies complex indirect reflections into a differentiable three-segment propagation model, as illustrated by the distinct green and purple paths originating from different scattering points on the human subject.}
    \label{fig:combined}
\end{figure}

\subsection{Indirect Path Simulation: A Unified Multi-Scatterer Surrogate}
\label{sec:indirect_path}
The indirect path is exclusively dedicated to multi-bounce reflections ($\ge 2$ bounces), encompassing human-to-wall, floor-to-wall, and wall-to-wall interactions. To strictly prevent double-counting with the single-bounce direct path, explicit multi-bounce ray tracing suffers from exponential computational complexity and is notoriously optimization-unfriendly. Instead of tracking unpredictable discrete rays, we introduce a physically defensible unified multi-scatterer formulation paired with proxy boundary-geometry.

Rather than treating environmental boundaries merely as smooth mirrors, we discretize both the dynamic human mesh and the static room layout (walls, floor, and ceiling) into a unified set of active scattering elements using geometrically constrained 3DGS. Let $\mathcal{P}_{ind} = \mathcal{P}_{human} \cup \mathcal{P}_{env}$ denote this combined set of global scatterers for the indirect path. 

Let $\mathcal{K}$ denote the set of room boundaries. We instantiate a set of proxy virtual receivers $\mathbf{x}_{w,k}$ on each boundary $k \in \mathcal{K}$. This design converts the complex, intractable multi-bounce integral into an efficient, differentiable three-segment propagation model: transmitter $\mathbf{x}_{\text{tx}} \rightarrow$ intermediate scatterer $p \in \mathcal{P}_{ind} \rightarrow$ proxy receiver $\mathbf{x}_{w,k} \rightarrow$ physical receiver $\mathbf{x}_{rx}$.

Because the intermediate scatterer $p$ is sampled from the unified set $\mathcal{P}_{ind}$, this formulation acts as a physically-constrained neural representation that explicitly learns the site-specific environmental multipath signatures. Rather than analytically calculating secondary bounces, it encodes the spatial multipath texture into the trainable parameters of the 3DGS. The 3DGS decoders will predict parameters like spherical harmonics, opacity, scaling, and offset for each patch~\cite{10.1145/3592433}, yielding a directional amplitude $SH_{p,k}(\mathbf{x}_p(t), \boldsymbol{\omega}_{p \rightarrow w,k})$ evaluated toward the proxy receiver $\mathbf{x}_{w,k}$. To prevent the volumetric inflation typical of unconstrained Gaussians, these scattering elements are formulated as flattened 2D surface patches that strictly adhere to the underlying geometric topology.

The global indirect signal is synthesized by coherently summing the contributions over all proxy boundaries and their respective visible scatterers (excluding self-intersections on the same boundary):
\begin{equation}
Sim_{ind}(t) = \sum_{k \in \mathcal{K}} \sum_{p \in vis(k)} \frac{SH_{p,k} \exp(j2\pi(f_0 \tau_{p,k}(t) + St\tau_{p,k}(t)))}{\|\mathbf{x}_{\text{tx}} - \mathbf{x}_p(t)\| \cdot \|\mathbf{x}_p(t) - \mathbf{x}_{w,k}\| \cdot \|\mathbf{x}_{w,k} - \mathbf{x}_{rx}\|}
\label{eq:indirect_path}
\end{equation}
where $\tau_{p,k}(t) = (\|\mathbf{x}_{\text{tx}} - \mathbf{x}_p(t)\| + \|\mathbf{x}_p(t) - \mathbf{x}_{w,k}\| + \|\mathbf{x}_{w,k} - \mathbf{x}_{rx}\|)/c$ defines the proxy propagation delay. 

By directly aggregating the multipath energy scattered from both the human body and the environment toward the proxy boundaries, this global summation acts as a robust, optimization-friendly surrogate for multi-bounce ray tracing, preserving complex multipath interference signatures by encoding them as spatial textures, entirely bypassing the prohibitive computational overhead of analytical multi-bounce calculations.

\subsection{Loss Function}
\label{sec:loss}
 To bridge the sim-to-real gap prior to frequency transformation, we inject zero-mean complex Gaussian noise into the idealized synthetic signal. The noise standard deviation is dynamically parameterized by a learnable rate multiplied by the signal's mean amplitude (further analyzed in Sec. \ref{sec:ablation2}).

Directly optimizing the raw time-domain complex signals is highly intractable. Radar absolute phase is extremely sensitive to sub-millimeter errors in kinematic tracking and geometric reconstruction; these inevitable sim-to-real misalignments completely distort the absolute phase, rendering raw complex Mean Squared Error (MSE) unreliable for network supervision~\cite{chen2023differentiableradiofrequencyray, richeek2024mmwave}. Therefore, we adopt an amplitude-only strategy, processing signals into Range-Doppler (RD) amplitude heatmaps via 2D Fast Fourier Transforms (FFT). 

Crucially, the proxy-geometry formulation delineates the interface between kinematics-driven path evolution and data-driven electromagnetic effects. While real-world multipath involves complex and often intractable phenomena (e.g., polarization, frequency-dependent absorption, and reflection-dependent phase offsets), our hybrid design assigns complementary roles to geometry and learning. The proxy delay $\tau_{p,k}(t)$ explicitly models the time-varying propagation delay, thereby enforcing that the dominant Doppler evolution is governed by the geometric rate of path-length change ($d\tau/dt$). Under this formulation, static, material-dependent phase offsets affect mainly the relative interference pattern but do not alter the kinematics-induced Doppler trend encoded by $\tau_{p,k}(t)$. In parallel, the neural 3DGS parameters ($SH_{p,k}$) serve as a compact surrogate to absorb unmodeled electromagnetic attenuation and scattering strength. This separation constrains the degrees of freedom of the learned component and empirically reduces the tendency to fit motion-irrelevant heatmap textures, as supported by our ablation study in Sec.~\ref{sec:ablation1}

Furthermore, raw radar reflections exhibit immense dynamic range, where strong primary reflections can monopolize gradients and overshadow vital kinematic details (more detail in Supplementary Material Sec.~\ref{sec:log_loss}). We balance this optimization penalty by applying a $20 \log_{10}(\cdot)$ transformation to convert amplitudes to a decibel (dB) scale.

The framework is optimized end-to-end by minimizing the Mean Squared Error (MSE) between the logarithmic heatmaps:
\begin{equation}
    \mathcal{L} = \frac{1}{N} \sum_{n=1}^{N} \left( 20 \log_{10}\left( \left| \mathcal{F}_{RD}(Sim) \right|_n \right) - 20 \log_{10}\left( \left| \mathcal{F}_{RD}(GT) \right|_n \right) \right)^2,
\end{equation}
where $\mathcal{F}_{RD}(\cdot)$ denotes the sequential 2D RD FFT operation, and index $n$ iterates over all $N$ bins in the flattened RD heatmap matrix.

\section{Experiments}
\subsection{Implementation Details}
We implement our framework using the mmMesh dataset~\cite{10.1145/3458864.3467679}, which collects signals via a commercial TI AWR1843BOOST mmWave radar equipped with 3 transmitting (TX) and 4 receiving (RX) antennas. For the training phase, we select eight distinct action sequences performed by a single primary subject. Each sequence comprises approximately 3,000 frames, and we allocate 80\% of this data to optimize the network weights. We train the entire framework end-to-end on a single NVIDIA RTX 4090 GPU, which requires a peak memory footprint of approximately 20 GB and takes roughly seven hours. During the inference phase, synthesizing an action takes under 1 second per frame.

\subsection{Baselines}
The selection of baselines is strictly constrained to approaches that share comparable hardware and signal configurations with the experimental setup. The synthesis of intermediate frequency (IF) and RD heatmaps is inherently coupled with the underlying radar signal model, which encompasses the waveform, center frequency, bandwidth, and chirp schedule, alongside the antenna array configuration, such as transceiver geometry and the arrangement of virtual channels. Many existing mmWave synthesis methods assume divergent transceiver setups or rely on multi-view sensing, rendering the corresponding outputs fundamentally incomparable to the proposed single-radar pipeline. Crucially, the closed-source nature of several contemporary approaches precludes exact reproduction and fair benchmarking. Consequently, the comparative evaluation primarily focuses on mmGPE~\cite{10.1145/3570361.3613302} and RF-Genesis~\cite{10.1145/3625687.3625798}, as these frameworks most closely align with the targeted configuration.

\subsection{Evaluation Protocol}

To assess synthesis quality and cross-subject generalization under a fixed scene, we establish 2 testing settings. For intra-subject synthesis, we evaluate the training subject's remaining 20\% frames containing unseen dynamic poses. For cross-subject evaluation, we test the model on a new human subject performing the same eight action categories, each comprising 3,000 frames. This confirms whether the framework overfits the primary subject's body shape or learns transferable physical scattering properties. We quantitatively measure generated RD heatmap fidelity using 3 standard metrics: Peak Signal-to-Noise Ratio (PSNR), Structural Similarity Index (SSIM), and Learned Perceptual Image Patch Similarity (LPIPS). All experiments are conducted in the same room layout, and more details about the dataset composition are provided in Supplementary Material Sec.~\ref{sec:dataset}.

\subsection{Main Results: Range-Doppler Synthesis}

\begin{figure}[t]
    \centering
    \includegraphics[width=1\linewidth]{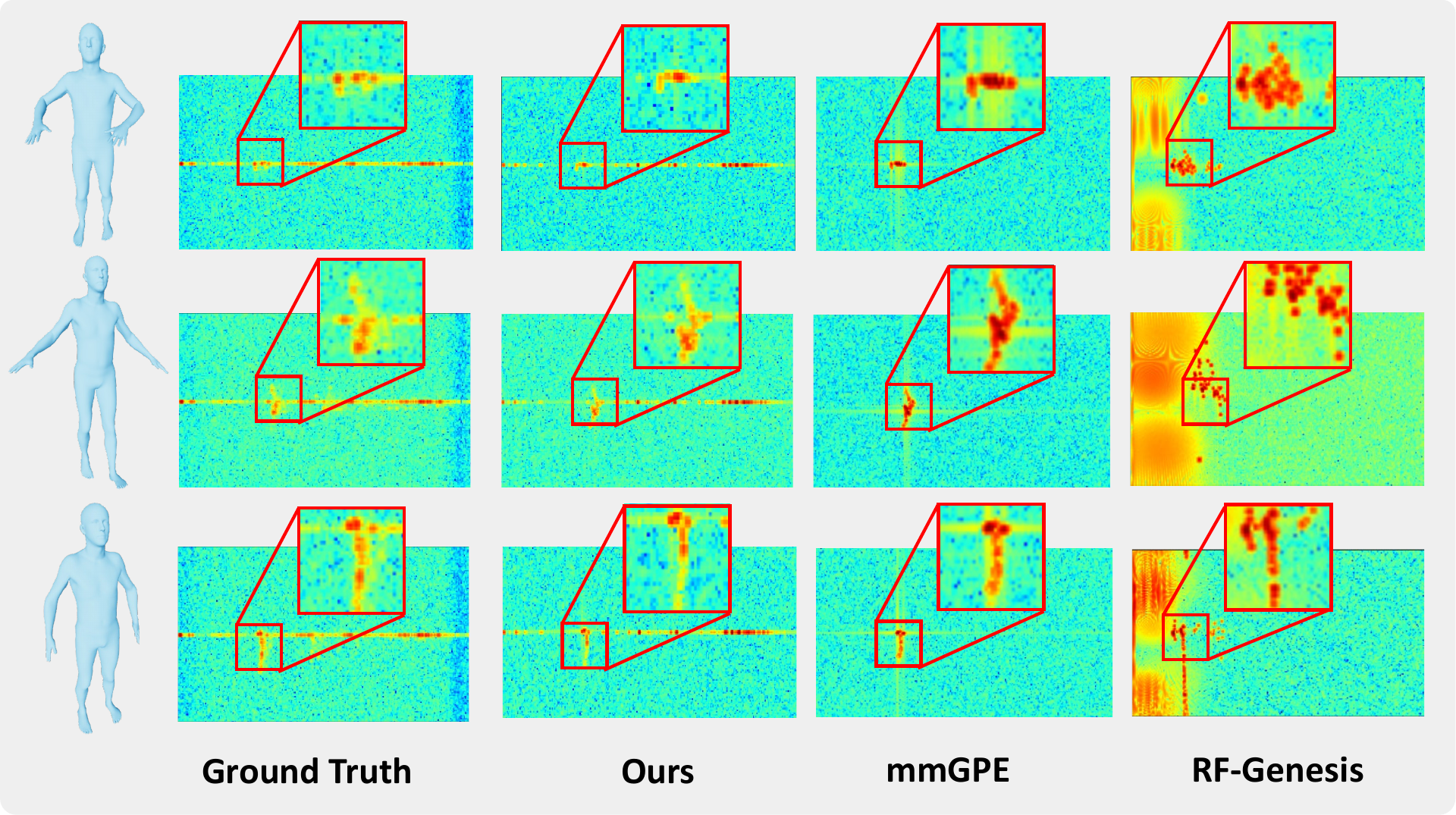}
    \caption{\textbf{Qualitative comparison of RD heatmaps.} The red boxes provide zoomed-in views of the micro-Doppler signatures generated by articulated human limb movements. Compared to existing baseline simulators (mmGPE and RF-Genesis), HybridSim synthesizes signals that exhibit significantly higher visual fidelity to the physical ground truth, accurately preserving fine-grained kinematic details.}
    \label{fig:Qualitative_Results}
\end{figure}

As shown in Table~\ref{tab:main_results}, HybridSim outperforms all baselines in both settings, notably achieving a significant LPIPS reduction that indicates superior structural fidelity. This is visually corroborated in Fig.~\ref{fig:Qualitative_Results}: our framework accurately preserves fine-grained kinematics matching the physical ground truth. Additional qualitative decomposition of the direct path, indirect path, and learned noise contributions is provided in Supplementary Material Sec.~\ref{sec:path_visualization}.

\input{table/main_psnr}
\input{table/HAR}

\subsection{Downstream Task: Human Activity Recognition}
\label{sec:HAR}
To validate the practical utility of the proposed simulator in bridging the sim-to-real domain gap, we evaluate the performance of the framework on a downstream human activity recognition (HAR) task. We employ the mmAP model~\cite{10825019} as the classification backbone, which utilizes three types of heatmaps (\ie, time-Doppler, time-range, and time-angle) to categorize human activities. 


Under a sim-to-real evaluation protocol, we train the classifier exclusively on synthetic data generated for an unseen subject across eight action categories, and then test it directly on corresponding physical ground truth measurements. The inclusion of this downstream evaluation is critical, as standard reconstruction metrics often fail to fully reflect the fidelity of radar cross-section (RCS) dynamics. Metrics like PSNR provide a global measure of pixel-level similarity but suffer from inherent bias due to the sparsity of radar data. Since micro-Doppler signatures of human motion occupy only a marginal region within the heatmaps, the background noise floor dominates the spatial area. Thus, substantial improvements in synthesizing critical kinematic semantics yield only incremental gains in global PSNR. 

The downstream HAR task provides a more representative assessment by heavily penalizing physically unrealistic artifacts and missing micro-Doppler components. As Table~\ref{tab:har_results} shows, the model trained on HybridSim data achieves a recognition accuracy of $92.07\%$. This constitutes an absolute gain of $37.85\%$ over mmGPE ($54.22\%$), even though HybridSim shows only a modest $10.9\%$ improvement in PSNR (from $20.09$ to $22.29$) under the cross-subject setting (Table~\ref{tab:main_results}). This discrepancy confirms that classification accuracy on real sensor data serves as a more reliable indicator of whether a simulator preserves the essential motion semantics required for robust real-world radar perception.

To further evaluate cross-subject generalization and increase the diversity of the downstream assessment, we additionally conduct HAR evaluation on another held-out human subject. These supplementary results are reported in Supplementary Material Sec.~\ref{sec:add_eval} and further confirm the robustness of HybridSim. 


\input{table/ablation_psnr}
\subsection{Ablation Study: Impact of Decoupled Hybrid Rendering}
\label{sec:ablation1}
To evaluate the necessity of our decoupled physics-learning architecture, we conduct an ablation study focusing on the rendering formulation. We construct a variant termed Coupled 3DGS, where the direct and indirect signal paths are unified and modeled entirely using Gaussian Patches. This configuration removes the BRDF model, forcing the network to implicitly learn all wave-surface interactions using Spherical Harmonics. Consistent with our cross-subject protocol, we evaluate this variant on the RD heatmap synthesis task and the downstream human activity recognition task using the unseen subject.

Quantitative results, summarized in Table~\ref{tab:ablation_coupled}, show a performance degradation across all metrics when the paths are coupled. Specifically, the PSNR drops significantly from 22.29 dB to 19.12 dB. Because the Gaussian patch relies on low-degree Spherical Harmonics to represent view-dependent amplitude, it struggles to accurately synthesize the sharp, highly directional specular reflections characteristic of the primary direct path. Furthermore, unifying the direct and indirect paths within a single representation forces the network to implicitly resolve complex multipath interference, severely entangling the dynamic human kinematics with the static environmental reflections. This lack of physical decoupling and explicit BRDF constraints heavily impacts downstream performance, with classification accuracy falling from 92.07\% to 85.61\% and the F1-score decreasing from 0.9216 to 0.8492 compared to our full HybridSim model. This decline confirms that when these distinct physical phenomena are coupled, the network overfits to the training data and fails to capture the intrinsic micro-Doppler semantics required for robust sim-to-real generalization. These findings substantiate that explicitly disentangling primary surface reflections from complex environmental multipath effects is essential for high-fidelity radar simulation.

\subsection{Ablation Study: Contribution of Direct and Indirect path}
Fig.~\ref{fig:path_decomposition} in the suppl. shows that the direct path carries the dominant human-motion response, while the indirect path contributes weaker scene-dependent multipath reflections. Quantitatively, removing the indirect path reduces HAR performance from 92.07\% accuracy / 0.9216 F1-score to 88.96\% / 0.8874. This shows that the indirect path is not the main source of human-motion cues, but provides complementary multipath information that improves Sim2Real alignment.

We also add a two-bounce geometric ray-tracing baseline based on ideal specular reflection. In this baseline, the direct path is kept unchanged as ours, and only the 3DGS-based indirect path is replaced. This baseline requires 3.44s/frame, whereas our method requires under 1s/frame, reducing the per-frame runtime by over 70\%.. And its HAR accuracy drops to 41.6\%, mainly because fixed empirical reflection coefficients cannot model site-specific multipath responses. The resulting spurious high-energy reflections distort RD heatmaps and hurt HAR training, supporting our indirect-path design.

\subsection{Ablation Study: Site-specific Second-scene Adaptation.}
Since the mmMesh dataset does not contain a second room layout, we repurpose another self-collected dataset. It contains one male with 12 new actions, 8 actions for training and 4 for evaluation. These actions have smaller motion amplitudes than those in main paper (details in Suppl. Sec.~\ref{sec:dataset}). HybridSim achieves 22.01 dB PSNR, 0.241 SSIM, 0.108 LPIPS, and 88.68\% accuracy / 0.8875 F1-score for HAR. This shows that HybridSim can be adapted to a new room layout. The lower HAR here is partly due to weaker limb-related reflections caused by smaller motion amplitudes, which makes fine-grained action discrimination more difficult.

\subsection{Ablation Study: Effect of Noise Floor on Sim-to-Real Alignment}
\label{sec:ablation2}

To evaluate the impact of dynamic noise injection on bridging the sim-to-real domain gap, we compare HybridSim variants with different noise-floor strategies under the HAR task. As introduced in Sec.~\ref{sec:loss}, idealized physical simulations inherently lack the sensor noise present in real-world measurements. We delegate the approximation of the unpredictable environmental and hardware noise floor to statistical learning. Without injecting synthetic noise, a downstream classifier trained on pristine simulated data tends to overfit to an unrealistically clean background. As shown in Table~\ref{tab:ablation_noise}, the variant without any noise injection, denoted as HybridSim (w/o noise), achieves an accuracy of only 72.79\% and an F1-score of 0.6790. This significant performance drop highlights the necessity of background noise for robust sim-to-real transfer.

\begin{figure}[t]

    \centering

    \includegraphics[width=1\linewidth]{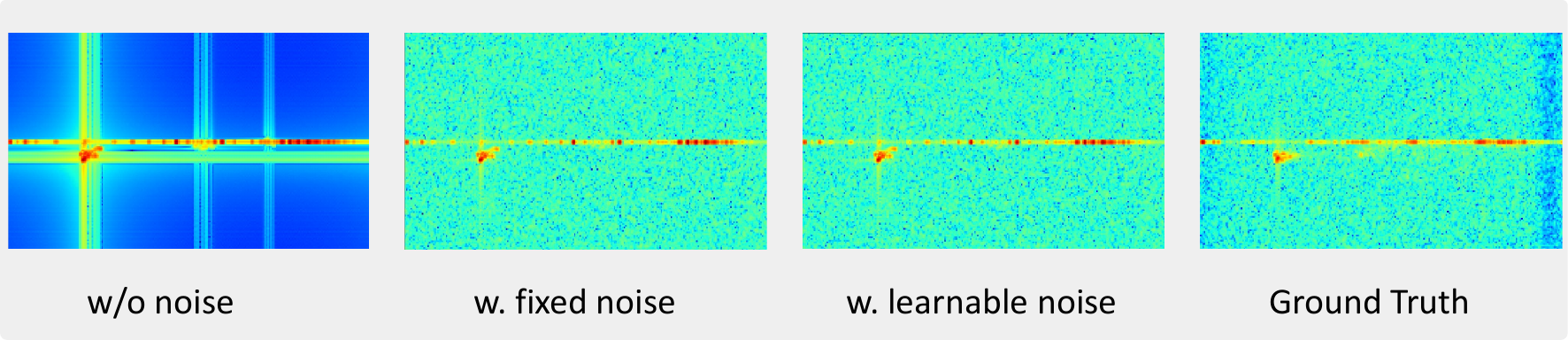}

    \caption{\textbf{Visual comparison of Range-Doppler heatmaps generated under different noise injection strategies.} From left to right: simulation without noise, simulation with a fixed noise, simulation with a learnable noise, and ground truth. The learnable noise strategy produces a background noise distribution that most closely resembles the real-world sensor measurements.}

    \label{fig:ablation_noise}
\end{figure}

To establish a control variable for comparison, we evaluate a variant injected with a fixed noise floor. It uses the same constant noise level as the baseline methods in Sec.~\ref{sec:HAR}; details are provided in Supplementary Material Sec.~\ref{sec:fixed_noise}. Applying this matching static noise elevates the classification accuracy to 90.51\% and the F1-score to 0.9068. Surpassing the baselines under the exact same noise constraint proves that the primary performance driver is the decoupled physics-learning architecture rather than the noise augmentation. The full model with a dynamically optimized noise level achieves the highest accuracy of 92.07\% and an F1-score of 0.9216. This progressive sim-to-real alignment is visually evident in Fig.~\ref{fig:ablation_noise}. Compared to the unrealistically clean background of the zero-noise simulation, learning the optimal noise rate during training directly aligns the synthetic data distribution with the statistical properties of the physical ground truth, further closing the sim-to-real gap.

\input{table/ablation_noise}

\section{Conclusion \& Limitation}
In this paper, we present HybridSim, a physics-learning hybrid simulator designed to synthesize high-fidelity mmWave radar RD heatmaps from dynamic human meshes. By explicitly decoupling signal propagation into an inverse-rendering-based direct path and a 3DGS-based indirect path, the framework successfully disentangles primary surface scattering from complex environmental multipath effects. Extensive evaluations, alongside substantial improvements in downstream human activity recognition tasks, demonstrate the superior capability of HybridSim in bridging the sim-to-real domain gap compared to existing baselines.

While HybridSim exhibits robust cross-subject generalization for unseen human targets, the current static scene representation is optimized for specific training environments, requiring parameter fine-tuning for novel indoor layouts. A compelling direction for future research involves integrating generalized neural scene priors to decouple the environmental feature embeddings from specific room geometries. This extension aims to facilitate zero-shot or rapid few-shot adaptation to arbitrary novel environments, thereby further advancing the scalability of synthetic radar data generation for ubiquitous human sensing applications.


\section*{Acknowledgements}
This work was supported in part by the Xiamen University Malaysia Research Fund Cycle 14/2024 under Grant No. XMUMRF/2024-C14/IECE/0052.

%
%
\bibliographystyle{splncs04}
\bibliography{main}
\input{suppl}
\end{document}

%% file: table/main_psnr.tex
\begin{table}[t]
    \centering
    \caption{\textbf{Quantitative Evaluation on the mmMesh Dataset.} We report the PSNR, SSIM, and LPIPS metrics for the generated RD heatmaps. The evaluation is divided into two settings: Intra-Subject (testing on the 20\% unseen frames of the training subject) and Cross-Subject (testing on 8 actions each with full 3,000 frames of an unseen human in the same scene).}
    \resizebox{\linewidth}{!}{
    \begin{tabular}{lcccccc}
        \toprule
        \multirow{2}{*}{\textbf{Method}} & \multicolumn{3}{c}{\textbf{Unseen Frames (Intra-Subject)}} & \multicolumn{3}{c}{\textbf{Unseen Human (Cross-Subject)}} \\
        \cmidrule(lr){2-4} \cmidrule(lr){5-7}
        & \textbf{PSNR$\uparrow$} & \textbf{SSIM$\uparrow$} & \textbf{LPIPS$\downarrow$} & \textbf{PSNR$\uparrow$} & \textbf{SSIM$\uparrow$} & \textbf{LPIPS$\downarrow$} \\
        \midrule
        RF-Genesis~\cite{10.1145/3625687.3625798} & 20.03 & 0.219 & 0.274 & 20.02 & 0.218 & 0.271 \\
        mmGPE~\cite{10.1145/3570361.3613302} & 20.09 & 0.221 & 0.261 & 20.09 & 0.220 & 0.253  \\
        \textbf{HybridSim (Ours)} & \textbf{22.30} & \textbf{0.262} & \textbf{0.089} & \textbf{22.29} & \textbf{0.262} & \textbf{0.084} \\
        \bottomrule
    \end{tabular}
    }
    \label{tab:main_results}
\end{table}

%% file: table/HAR.tex
\begin{table}[t]
    \centering
    \caption{\textbf{Downstream Task Human Activity Recognition Performance.} The mmAP classification network is trained exclusively on synthetic data generated by different simulators and tested directly on real-world ground truth measurements. Higher accuracy on real data indicates a smaller sim-to-real domain gap.}
    \begin{tabular*}{\linewidth}{@{\extracolsep{\fill}} l c c @{}}
        \toprule
        \textbf{Training Source (Simulator)} & \textbf{Accuracy (\%)$\uparrow$} & \textbf{F1-Score$\uparrow$} \\
        \midrule
        RF-Genesis & 30.68 & 0.2584 \\
        mmGPE & 54.22 & 0.4990 \\
        \textbf{HybridSim (Ours)} & \textbf{92.07} & \textbf{0.9216} \\
        \bottomrule
    \end{tabular*}
    \label{tab:har_results}
\end{table}

%% file: table/ablation_psnr.tex
\begin{table}[t]
    \centering
    \caption{\textbf{Ablation Study on Rendering Formulation.} We compare our full decoupled architecture against a Coupled 3DGS variant where both the direct and indirect signal paths are unified under a pure Gaussian Patch representation without using BRDF. The evaluation is conducted on the unseen subject with 8 actions reporting both RD synthesis fidelity and HAR performance.}
    \begin{tabular*}{\linewidth}{@{\extracolsep{\fill}} l c c c c c @{}}
        \toprule
        \multirow{2}{*}{\textbf{Method}} & \multicolumn{3}{c}{\textbf{Synthesis Metrics}} & \multicolumn{2}{c}{\textbf{Downstream Task (HAR)}} \\
        \cmidrule(lr){2-4} \cmidrule(lr){5-6}
        & \textbf{PSNR$\uparrow$} & \textbf{SSIM$\uparrow$} & \textbf{LPIPS$\downarrow$} & \textbf{Accuracy (\%)$\uparrow$} & \textbf{F1-Score$\uparrow$} \\
        \midrule
        HybridSim (Coupled 3DGS) & 19.12 & 0.252 & 0.099 & 85.61 & 0.8492 \\
        \textbf{HybridSim (w. BRDF)} & \textbf{22.29} & \textbf{0.262} & \textbf{0.084} & \textbf{92.07} & \textbf{0.9216} \\
        \bottomrule
    \end{tabular*}
    \label{tab:ablation_coupled}
\end{table}

%% file: table/ablation_noise.tex
\begin{table}[t]
    \centering
    \caption{\textbf{Quantitative evaluation of noise injection strategies on the downstream HAR task.} We compare the classification accuracy and F1-score of models trained on sim data without noise, with a fixed noise, and with learnable noise.}
    \begin{tabular*}{\linewidth}{@{\extracolsep{\fill}} l c c @{}}
        \toprule
        \textbf{Training Source (Simulator)} & \textbf{Accuracy (\%)$\uparrow$} & \textbf{F1-Score$\uparrow$} \\
        \midrule
        HybridSim (w/o noise) & 72.79 & 0.6790 \\
        HybridSim (w. fixed noise)& 90.51 & 0.9068 \\
        \textbf{HybridSim (w. learnable noise)} & \textbf{92.07} & \textbf{0.9216} \\
        \bottomrule
    \end{tabular*}
    \label{tab:ablation_noise}
\end{table}

%% file: suppl.tex
\newpage
\begin{center}
{\Large \textbf{Supplementary Materials for HybridSim}}
\end{center}
\setcounter{section}{5}
\setcounter{figure}{6}
\setcounter{table}{4}

This supplementary material provides additional implementation details, controlled analyses, and qualitative evidence that complement the main paper. In particular, it clarifies several technical components of HybridSim, expands the supporting experiments behind the main design choices, and presents additional visualizations to help interpret the synthesized mmWave signals.

\paragraph{Structure.}
\begin{itemize}
    \item Sec.~\ref{sec:supp_visibility} details the visibility and occlusion handling strategy for both direct and indirect propagation paths.
    \item Sec.~\ref{sec:brdf_details} presents the BRDF formulation used for direct-path scattering.
    \item Sec.~\ref{sec:log_loss} explains the logarithmic loss design and its effect on compressing dynamic range while preserving weak micro-Doppler signatures.
    \item Sec.~\ref{sec:add_eval} provides additional evaluation results on an unseen subject to further demonstrate the generalization capabilities of HybridSim.
    \item Sec.~\ref{sec:gcn_ablation} provides additional ablations on the GCN regularizer, with emphasis on training stability and inference-time behavior.
    \item Sec.~\ref{sec:dataset} describes the adopted mmMesh action categories and the dataset composition used in our experiments.
    \item Sec.~\ref{sec:fixed_noise} clarifies the fixed-noise control setting used in our experiments.
    \item Sec.~\ref{sec:path_visualization} presents additional qualitative visualizations of the direct-path, indirect-path, and learned-noise contributions in the synthesized mmWave signal.
\end{itemize}

\section{Visibility and Occlusion Handling}
\label{sec:supp_visibility}

A physically accurate radar simulation inherently requires precise handling of self-occlusions, line-of-sight blockages, and mutual occlusions between the dynamic subject and the static environment. To achieve this efficiently without resorting to computationally prohibitive ray-mesh intersection tests, the proposed framework dynamically determines the visibility of scattering centers using a Hidden Point Removal (HPR) algorithm \cite{10.1016/j.cag.2010.03.002}. To address the distinct physical geometries of the direct and indirect propagation paths, the visibility computation is explicitly decoupled as follows.

\textbf{Direct Path Visibility.} For the direct (single-bounce) signal path, visibility must account for both self-occlusions of the human body and mutual occlusions between the subject and the room. Rather than evaluating the human and the environment separately, the HPR algorithm is applied globally to a unified point cloud comprising both the posed human mesh vertices and the static scene patches. 

The viewpoint for this HPR operation is set precisely at the physical radar transceiver's location. By projecting this unified point cloud from the sensor's perspective, the algorithm naturally and efficiently resolves complex mutual occlusions. The resulting subset of unobstructed scatterers forms the candidate set $\text{vis}$ utilized in Eq.~\ref{eq:sim_d} of the main text.

\textbf{Indirect Path Visibility and Double-Counting Prevention.} The indirect path simulates multi-bounce reflections by routing signals through intermediate scatterers toward proxy virtual receivers $\mathbf{x}_{w,k}$ instantiated on each room boundary $k$. Consequently, the visibility subset $vis(k)$ must be computed independently for each proxy boundary. 

For a given boundary $k$, the HPR algorithm is executed from the perspective of the virtual receiver $\mathbf{x}_{w,k}$ against the unified global point cloud. However, to strictly prevent unphysical self-intersections and energy double-counting, a geometric filtering step is enforced prior to the HPR operation: any scene scatterer $p$ that structurally resides on the same boundary $k$ is explicitly removed from the candidate set. This mathematical constraint ensures that a wall cannot reflect electromagnetic energy directly onto itself. By evaluating the filtered point cloud from the vantage point of $\mathbf{x}_{w,k}$, the framework robustly captures coherent multi-bounce paths (\eg, human-to-wall or floor-to-wall) while maintaining strict energy conservation and avoiding overlap with the direct rendering pass.

\section{Details of the Microfacet BRDF Formulation}
\label{sec:brdf_details}
As described in Sec.~\ref{sec:direct_path} of the main text, the scattering amplitude $L_p$ for the direct path is computed using an inverse rendering formulation based on a microfacet-based scattering term. 

This model integrates a Trowbridge-Reitz normal distribution function \cite{10.5555/2383847.2383874}, a Smith geometric shadowing function \cite{Heitz2014UnderstandingTM}, and complex Fresnel equations \cite{10892639}. For a generic primary scatterer $p \in \mathcal{P}_{dir}$ at time $t$, the scattering amplitude response is modeled by:

\begin{equation}
    L_p(\mathbf{x}_p(t), \boldsymbol{\omega}_{i,p}, \boldsymbol{\omega}_{o,p}) = f_r(\boldsymbol{\omega}_{i,p}, \boldsymbol{\omega}_{o,p} ; \eta, k, \alpha, w), \label{eq:brdf_eq}
\end{equation}

where $\mathbf{x}_p(t)$ denotes the position of scatterer $p$ at time $t$. For the human subject, this position is obtained by applying Linear Blend Skinning to the canonical vertex, whereas for the room boundaries it corresponds to the spatial coordinates of the parameterized wall patches. The function $f_r$ denotes an amplitude-only microfacet response term parameterized by the complex refractive index components $(\eta, k)$, the surface roughness $\alpha$, and a learned polarization-mixing weight $w$. Given the incident direction $\boldsymbol{\omega}_{i,p}$ from the transmitter and the outgoing direction $\boldsymbol{\omega}_{o,p}$ toward the receiver, it produces the direct-path scattering amplitude at time $t$.

\begin{figure}[h]
    \centering
    \includegraphics[width=1\linewidth]{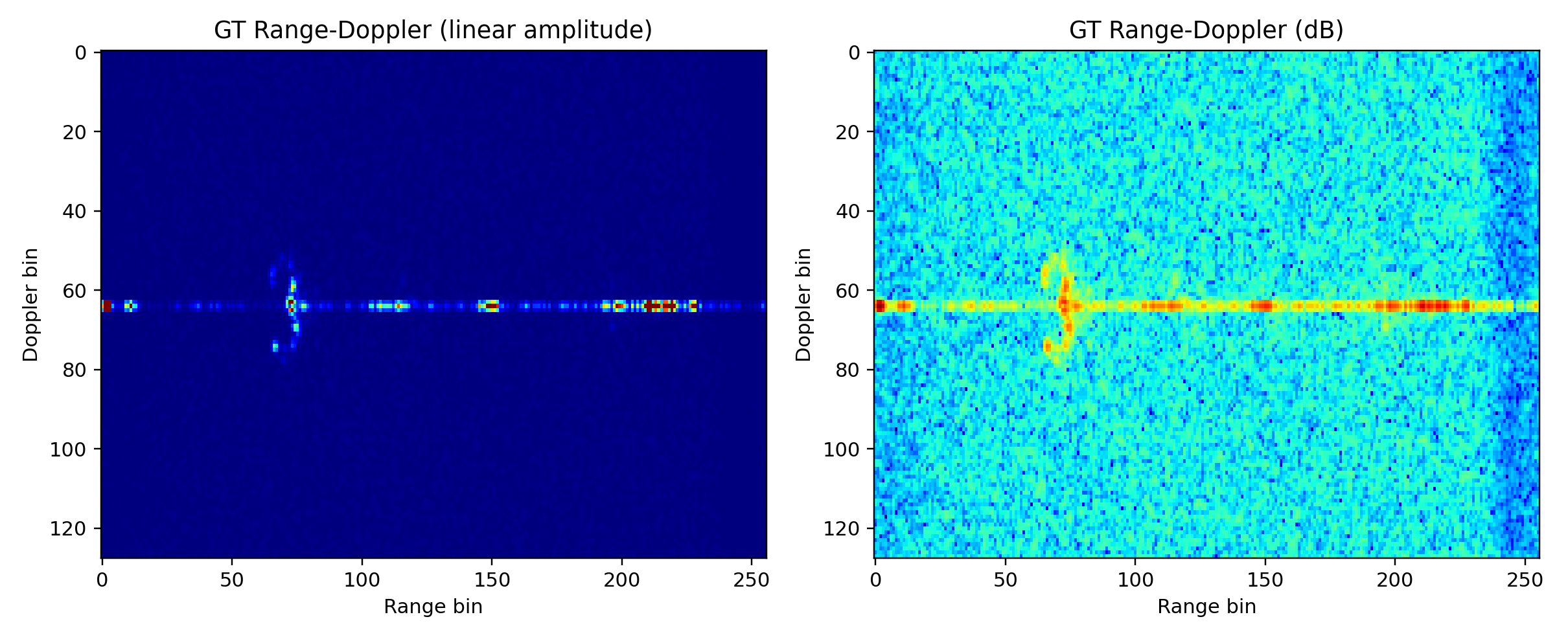}
    \caption{\textbf{Visual comparison of ground truth RD heatmaps in linear amplitude (left) and logarithmic decibel scale (right).} In the linear scale, the immense energy of primary stationary reflections dominates the signal, completely masking subtle kinematic movements. The logarithmic transformation effectively compresses the dynamic range, revealing the crucial micro-Doppler signatures generated by articulated limb motions.}
    \label{fig:linear_vs_db}
\end{figure}
\section{Logarithmic Transformation in Loss Function}
\label{sec:log_loss}
As outlined in Sec.~\ref{sec:loss} of the main text, the Loss function evaluates the MSE on a dB scale using a $20 \log_{10}(\cdot)$ transformation. Raw intermediate-frequency radar reflections exhibit an immense dynamic range. The energy returned by strong, stationary primary reflectors (such as the human torso or flat walls) can be exponentially larger than the subtle micro-Doppler signatures generated by articulated limbs. Computing the loss directly on linear amplitudes would cause these strong reflectors to monopolize the gradients during backpropagation, entirely overshadowing the semantically vital kinematic signatures, as visually demonstrated in Fig.~\ref{fig:linear_vs_db}. The logarithmic transformation acts as a crucial dynamic range compressor, evenly balancing the optimization penalty and ensuring that subtle micro-Doppler features are robustly learned.

\section{Additional HAR Evaluation}
\label{sec:add_eval}
We evaluate one additional unseen male subject in the original scene. HybridSim achieves 22.37 dB PSNR, 0.262 SSIM, 0.092 LPIPS, and 97.02\% accuracy / 0.9702 F1-score HAR performance. 


\section{Ablation Study: GCN as a Training-Time Regularizer}
\label{sec:gcn_ablation}
As discussed in Sec.~\ref{sec:direct_path} of the main text, the Graph Convolutional Network (GCN) is introduced not as an inference component, but as a training-time regularizer for stabilizing the inverse-rendering optimization on the articulated human mesh. Its role is to impose a weak spatial continuity prior during optimization, where the BRDF parameter estimation is otherwise highly ill-posed under sparse and noisy radar supervision. In this section, we provide additional ablations to clarify why the GCN is beneficial during training but should be deactivated at inference time.

\begin{figure}[h]
    \centering
    \includegraphics[width=0.95\linewidth]{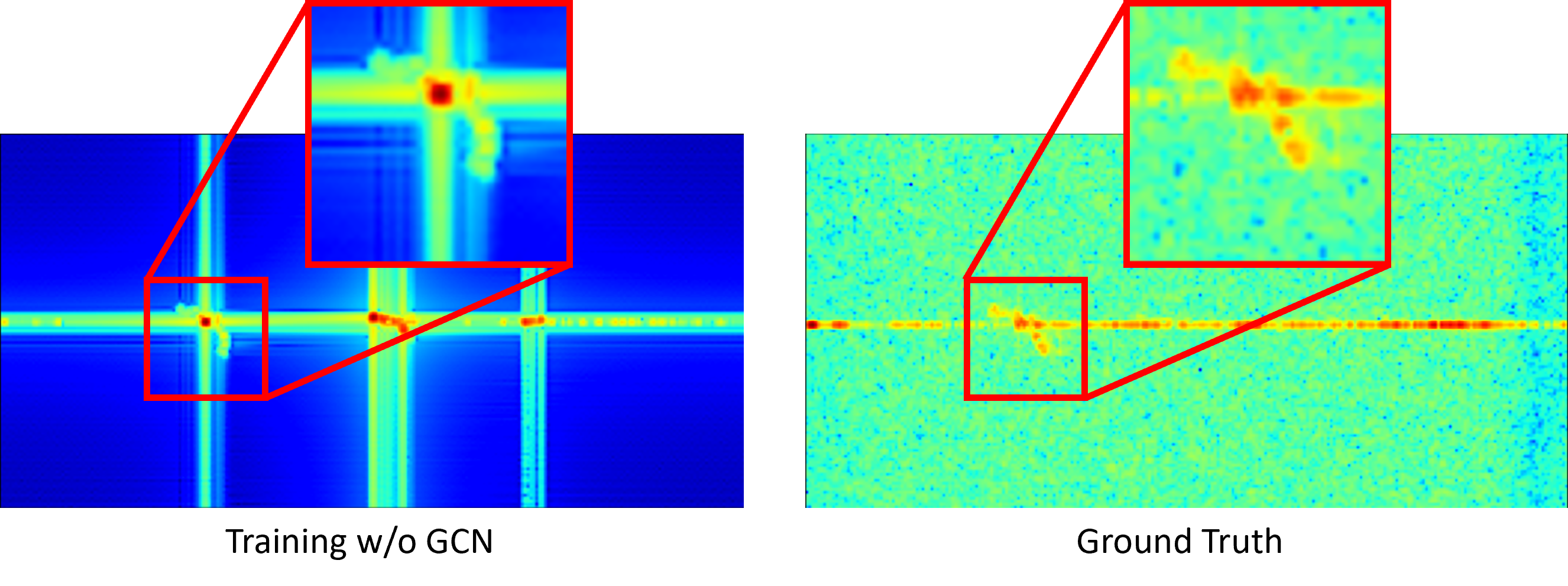}
    \caption{\textbf{Visual symptom of unstable training without GCN.} 
Without the GCN regularizer, the synthesized RD heatmap can exhibit an abnormal amplification of local body returns (red boxes), where a small subset of human-related scatterers becomes unrealistically dominant. This pathological behavior distorts the expected micro-Doppler pattern and is consistent with eventual optimization breakdown observed in late-stage training.}
    \label{fig:training_wo_gcn}
\end{figure}

\textbf{Training Stability.} Estimating the microfacet BRDF parameters (e.g., complex refractive index and roughness) from radar supervision is highly ill-conditioned, since the optimization is driven by sparse, noisy, and spatially non-uniform gradients on the human mesh. Without the GCN regularizer, these gradients tend to concentrate on a small subset of vertices, which leads to unstable updates and eventually causes loss spikes and parameter explosion (typically around epoch 7400 in our runs). This instability is also visible in the synthesized RD maps: as shown in Fig.~\ref{fig:training_wo_gcn}, the model trained without GCN can produce physically implausible, abnormally strong body reflections that locally dominate the RD heatmap and distort the expected motion structure. Such pathological amplification is consistent with the optimization breakdown observed in late-stage training. By introducing graph-based feature propagation together with stochastic masking during training, the GCN regularizes these localized updates and improves optimization stability, resulting in consistent convergence across runs.

\textbf{Inference Performance.} While the GCN is beneficial for stabilizing training, we deliberately bypass it during the inference phase. This design is intentional: the continuity prior is used to regularize the latent field during optimization, and applying graph propagation again at test time would introduce an additional low-pass bias that oversmooths fine-grained scattering variations. To examine this effect, we evaluate a variant where the unmasked GCN remains active during inference and compare it against our default inference strategy (without GCN) on both RD synthesis and the downstream HAR task.

As shown in Table~\ref{tab:ablation_study}, keeping the GCN active during inference leads to a clear drop in downstream HAR performance, with the overall accuracy decreasing from 92.07\% to 86.79\% and the Macro F1-Score from 0.9216 to 0.8629. In contrast, the global RD synthesis metrics remain nearly unchanged. This indicates that the adverse effect of GCN-enabled inference is not reflected primarily in the overall reconstruction quality of the full RD map, but rather in localized motion-sensitive regions that are critical for downstream recognition.

To further isolate the failure mode, we additionally evaluate Action 6 (\textit{walking back and forth with arm swing}) using static-removed RD heatmaps, so that the comparison focuses more on dynamic human returns rather than dominant static components. As shown in Table~\ref{tab:gcn_a6}, the difference becomes more visible on this motion-sensitive action: disabling the GCN during inference consistently improves all three synthesis metrics. This observation is consistent with the HAR degradation in Table~\ref{tab:ablation_study}, since Action 6 relies heavily on preserving weak limb-induced micro-Doppler signatures.
\input{table/ablation_gcn}
\input{table/ablation_gcn_psnr_a6}
\begin{figure}[ht]
    \centering

    \begin{minipage}{0.6\textwidth}
        \centering
        \includegraphics[height=3.8cm]{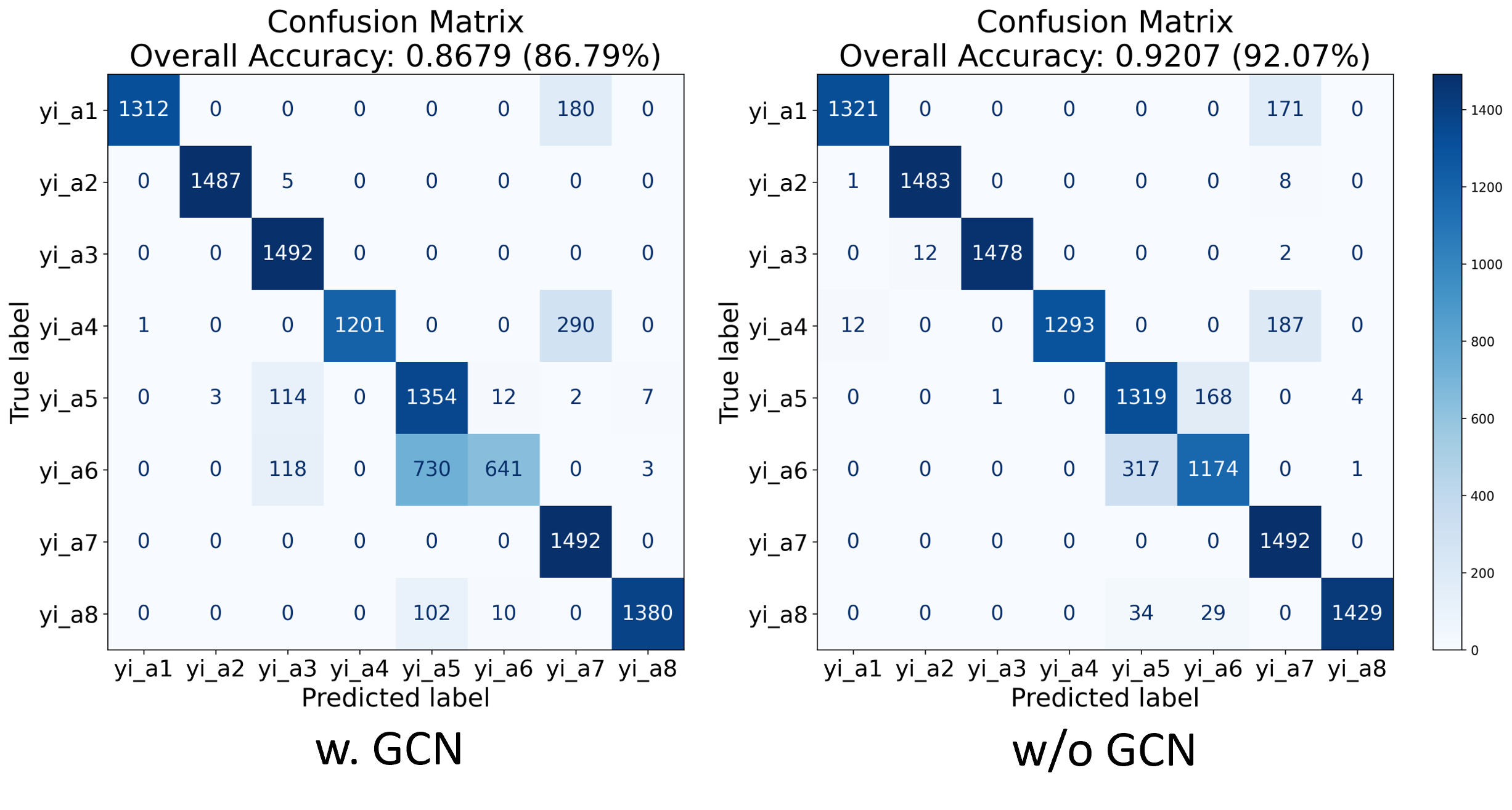}
        \vspace{0.15cm}
        \centerline{(a) Confusion Matrix Comparison}
    \end{minipage}\hfill
    \begin{minipage}{0.39\textwidth}
        \centering
        \includegraphics[height=3.8cm]{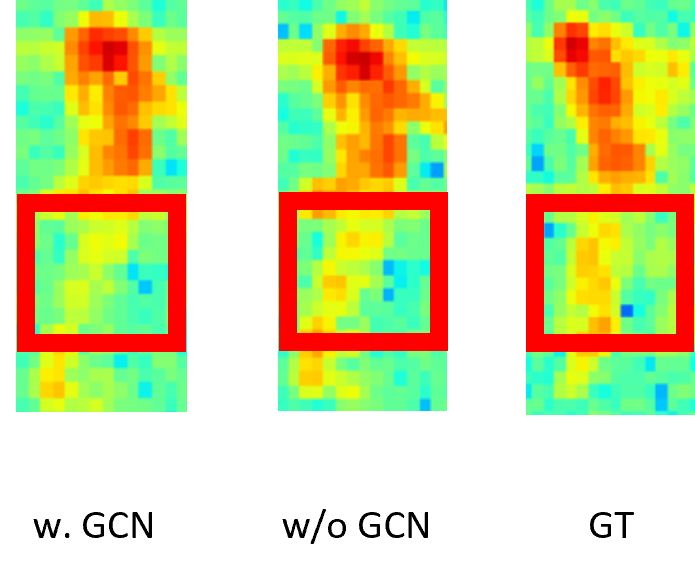}
        \vspace{0.15cm}
        \centerline{(b) RD Heatmap Comparison}
    \end{minipage}
    \caption{\textbf{Visual analysis of the effect of GCN-enabled inference.} (a) The confusion matrix shows increased confusion between highly similar activities when GCN remains active at test time, particularly between Action 6 and Action 5. (b) The Range-Doppler heatmaps suggest that GCN-enabled inference suppresses the responses associated with subtle limb motions, weakening the micro-Doppler signatures induced by arm swings.}
    \label{fig:gcn_visual}
\end{figure}

\textbf{Qualitative Analysis.} We hypothesize that the performance drop with GCN-enabled inference is mainly caused by excessive spatial smoothing, which suppresses legitimate high-frequency scattering variations that are critical for resolving fine-grained limb motions. Importantly, such degradation can be diluted in global image-level metrics, since PSNR/SSIM/LPIPS are averaged over the full RD map and are relatively insensitive to localized attenuation in limb-dominant micro-Doppler regions.

As illustrated in the confusion matrix comparison (Fig.~\ref{fig:gcn_visual}(a)), the model with GCN active frequently misclassifies Action 6 (\textit{walking back and forth with arm swing}) as Action 5 (\textit{walking back and forth}). Because these two activities share highly similar macroscopic torso trajectories, distinguishing them relies on the subtle micro-Doppler signatures generated by the arm swings.

As shown in the RD heatmap comparison (Fig.~\ref{fig:gcn_visual}(b)), applying the GCN during inference overly smooths the latent material embeddings, resulting in attenuated micro-Doppler responses for limb movements. This suggests that the continuity prior, while useful during optimization, becomes overly restrictive at test time and effectively double-counts smoothing on the learned latent field. By deactivating the GCN during inference, our framework better preserves these fine-grained kinematic cues, leading to more accurate classification.

\section{Dataset Details in main experiments}
\label{sec:dataset}
To evaluate the synthesis fidelity and validate the downstream utility of HybridSim, we utilize sequences from the mmMesh dataset~\cite{10.1145/3458864.3467679}, encompassing a diverse set of articulated human motions. Specifically, the dataset incorporates 8 distinct action categories that challenge the simulator's ability to model both macroscopic trajectories and fine-grained micro-Doppler signatures. The 8 actions are detailed as follows:

\begin{enumerate}
    \item \textbf{Torso rotations}: The subject stands in place and continuously twists their upper body.
    \item \textbf{Clockwise walking}: The subject walks in a continuous circular trajectory in a clockwise direction.
    \item \textbf{Counter-clockwise walking}: The subject walks in a continuous circular trajectory in a counter-clockwise direction.
    \item \textbf{Arm swing}: The subject stands still while randomly swinging his/her arms horizontally, upward, or downward.
    \item \textbf{Walking back and forth}: The subject walks straight toward the radar and then walks backward, keeping the arms relatively still.
    \item \textbf{Walking back and forth with arm swing}: The subject walks straight toward the radar and backward while actively and naturally swinging their arms.
    \item \textbf{Walking in the place}: The subject performs a marching motion without spatial displacement.
    \item \textbf{Lunges}: The subject keeps performing a lunge pose, alternating between using his/her left and right leg.
\end{enumerate}

\noindent This diverse action space ensures that the generated mmWave signals are rigorously tested against a wide spectrum of kinematic complexities, ranging from dominant torso reflections to highly dynamic, low-energy limb scatterings.

\section{Details of the Fixed-Noise Control Setting}
\label{sec:fixed_noise}

As described in Sec.~\ref{sec:ablation2} of the main text, we include a fixed-noise variant as a control setting to separate the effect of background-noise injection from that of the proposed rendering architecture. In this setting, the noise level is not learned during training. Instead, we inject zero-mean complex Gaussian noise with the same fixed rate for all samples and all compared simulators.

Let $Sim(t)$ denote the synthesized complex intermediate-frequency signal before noise injection, and let
\[
\mu_{Sim} = \mathrm{mean}(|Sim(t)|).
\]
The standard deviation of the injected noise is defined as
\[
\sigma_{\mathrm{fix}} = r_{\mathrm{fix}} \cdot \mu_{Sim},
\]
where $r_{\mathrm{fix}} = 0.025$ is a fixed scalar chosen from empirical dataset statistics as a coarse global approximation of the background noise level. The resulting noisy signal is
\[
Sim_{\mathrm{fix}}(t) = Sim(t) + \sigma_{\mathrm{fix}}\left(\epsilon_{\mathrm{real}} + j\,\epsilon_{\mathrm{imag}}\right),
\]
where $\epsilon_{\mathrm{real}}, \epsilon_{\mathrm{imag}} \sim \mathcal{N}(0,1)$ are i.i.d. standard Gaussian variables.

The key point is that $r_{\mathrm{fix}}$ is shared across all methods and is not re-tuned per action, subject, or simulator. Therefore, this setting serves as a strict control experiment: any performance difference under fixed noise should mainly reflect the rendering architecture itself, rather than method-specific noise optimization. In contrast, the full model in the main text replaces this fixed rate with a learnable one to better match the background statistics of real measurements.

\section{More Qualitative Results on Direct and Indirect Path Contributions}
\label{sec:path_visualization}

\begin{figure*}[h]
    \centering
    \includegraphics[width=0.99\textwidth]{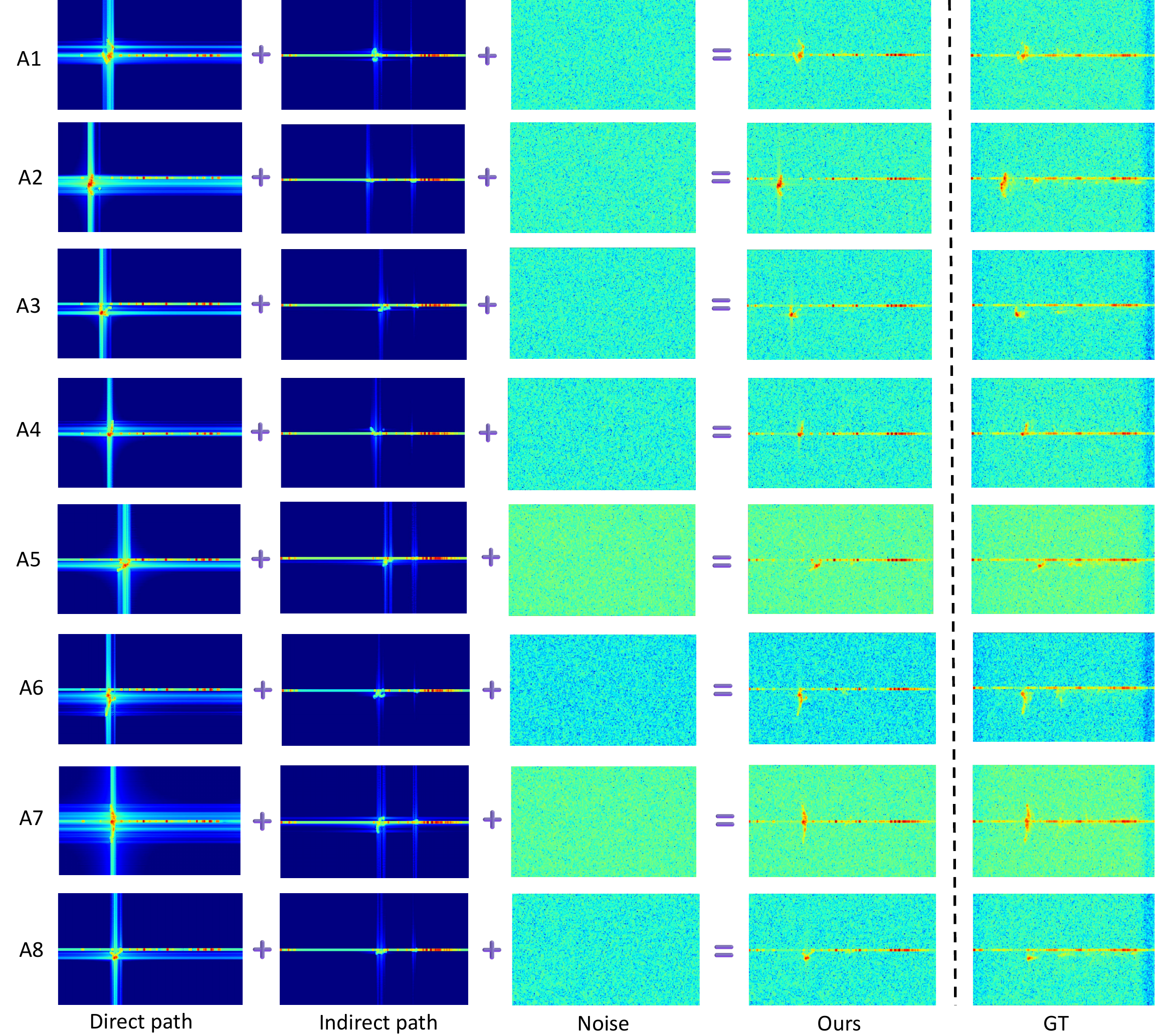}
    \caption{Qualitative decomposition of HybridSim in RD space. For the same motion frame or short sequence, we show the direct-path signal, the indirect-path signal, the learned noise term, the final HybridSim output, and the corresponding ground truth. The direct path explains the dominant target return and the main micro-Doppler backbone; the indirect path adds weaker but structured room-dependent multipath responses; the learned noise term restores realistic background statistics. The final result is obtained by composing all three components. Action labels A1--A8 follow the action definitions in Sec.~10.}
    \label{fig:path_decomposition}
\end{figure*}

Fig.~\ref{fig:path_decomposition} visualizes how the three components of HybridSim contribute to the final mmWave signal. The direct-path branch captures the dominant target return and most of the micro-Doppler structure. The indirect-path branch contributes weaker but spatially structured responses caused by room-dependent multipath interactions. The learned noise term restores the non-zero background statistics that are missing from the deterministic path components alone.

The key message of this figure is that these three components play different and complementary roles. The direct path provides the main signal backbone, the indirect path supplements structured environmental multipath, and the learned noise term improves realism in the background distribution. Only after combining all three do the synthesized RD maps better match the ground-truth measurements.


%% file: table/ablation_gcn.tex
\begin{table}[t]
    \centering
    \caption{\textbf{Quantitative ablation of the GCN module during the inference phase.} The evaluation is conducted on the unseen subject with 8 actions, reporting both RD synthesis fidelity and HAR performance. Disabling the GCN during inference yields the best downstream performance, while the global synthesis metrics remain nearly unchanged.}
    \begin{tabular*}{\linewidth}{@{\extracolsep{\fill}} l c c c c c @{}}
        \toprule
        \multirow{2}{*}{\textbf{Inference Setting}} & \multicolumn{3}{c}{\textbf{Synthesis Metrics}} & \multicolumn{2}{c}{\textbf{Downstream Task (HAR)}} \\
        \cmidrule(lr){2-4} \cmidrule(lr){5-6}
        & \textbf{PSNR$\uparrow$} & \textbf{SSIM$\uparrow$} & \textbf{LPIPS$\downarrow$} & \textbf{Accuracy (\%)$\uparrow$} & \textbf{F1-Score$\uparrow$} \\
        \midrule
        w. unmasked GCN & 22.292 & 0.2623 & 0.085 & 86.79 & 0.8629 \\
        \textbf{w/o GCN (Ours)} & \textbf{22.294} & \textbf{0.2623} & \textbf{0.084} & \textbf{92.07} & \textbf{0.9216} \\
        \bottomrule
    \end{tabular*}
    \label{tab:ablation_study}
\end{table}

%% file: table/ablation_gcn_psnr_a6.tex
\begin{table}[t]
    \centering
    \caption{\textbf{Activity-specific synthesis comparison on Action 6 (walking back and forth with arm swing).} To better isolate the dynamic human return, the comparison is performed on static-removed RD heatmaps. Unlike the nearly unchanged global metrics in Table~5, the effect of GCN-enabled inference becomes more visible on this motion-sensitive action.}
    \label{tab:gcn_a6}
    \begin{tabular*}{0.72\linewidth}{@{\extracolsep{\fill}} l c c c @{}}
        \toprule
        \textbf{Inference Setting} & \textbf{PSNR$\uparrow$} & \textbf{SSIM$\uparrow$} & \textbf{LPIPS$\downarrow$} \\
        \midrule
        w. unmasked GCN & 22.08 & 0.3017 & 0.0856 \\
        \textbf{w/o GCN (Ours)} & \textbf{22.41} & \textbf{0.3023} & \textbf{0.0854} \\
        \bottomrule
    \end{tabular*}
\end{table}